\documentclass[sigconf]{acmart}
\AtBeginDocument{%
  }
\usepackage{multirow}
\usepackage{adjustbox}
\usepackage{subcaption}
\usepackage{arydshln}

\setcopyright{acmlicensed}
\copyrightyear{2018}
\acmYear{2018}
\acmDOI{XXXXXXX.XXXXXXX}
\acmConference[Conference acronym 'XX]{Make sure to enter the correct
  conference title from your rights confirmation email}{June 03--05,
  2018}{Woodstock, NY}
\acmISBN{978-1-4503-XXXX-X/2018/06}

\renewcommand\footnotetextcopyrightpermission[1]{}  
\begin{document}

\title{Lossless Compression of Large Language Model-Generated Text via Next-Token Prediction}

\author{Yu Mao}
\affiliation{%
  \institution{MBZUAI}
  \streetaddress{P.O. Box 1212}
  \country{Abu Dhabi}
}
\email{yu.mao@mbzuai.ac.ae}

\author{Holger Pirk}
\affiliation{%
  \institution{Imperial College London}
  \country{England}
}
\email{hlgr@ic.ac.uk}

\author{Chun Jason Xue}
\affiliation{%
  \institution{MBZUAI}
  \country{Abu Dhabi}
}
\email{jason.xue@mbzuai.ac.ae}


\renewcommand{\shortauthors}{Trovato et al.}

\begin{abstract}
As the deployment and utilization of LLMs continue to expand, the volume of LLM-generated data is increasing at an unprecedented rate.
Therefore, the need for \textit{effective and lossless compression of LLM-generated data has become increasingly critical in text data management}.
However, compressing LLM-generated data poses significant challenges compared to traditional text data, such as human-generated or machine-generated content.
Traditional machine-gene\\rated data is predominantly derived from computational processes or device outputs, characterized by its highly structured and discrete nature, often limited to low-level information such as labels or numerical values. As a result, conventional lossless compression algorithms are able to exploit this structure to efficiently compress such data.
In contrast, the complexity and diversity of LLM-generated data require different approaches to achieve effective lossless compression.
To this end, this work conducts the first investigation into effective lossless compression methods for LLM-generated data.
Interestingly, since LLMs are trained through next-token prediction, we observe that LLM-generated data exhibits a high predictability for LLMs, which allows LLMs to achieve remarkable lossless compression of LLM-generated data.
We conduct extensive experiments with 14 representative LLMs and 8 LLM-generated datasets from different categories.
The experimental results reveal that LLM-based prediction methods achieve exceptional compression rates, exceeding 20×, which significantly outperforms the 3x achieved by the state-of-the-art lossless compressor Gzip.
Moreover, we demonstrate that the superior performance of LLM-based compressors generalizes across different LLM scales and diverse dataset categories, making them a useful tool in lossless data compression.
\end{abstract}

\maketitle

\section{Introduction}

\begin{figure}[t]
    \vspace{0.15in}
    \centering
    \includegraphics[width=0.9\linewidth]{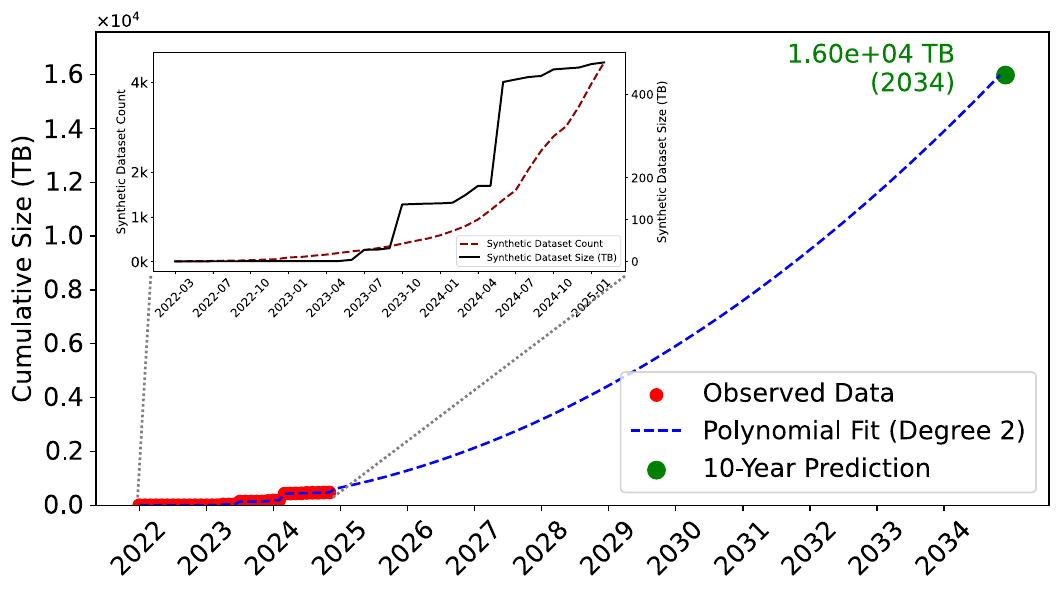} 
    \caption{Trend of increasing LLM-generated data.}
    \label{fig:trend} 
\end{figure}
The proliferation of large language models (LLMs) has led to a significant surge in AI-generated text across diverse domains~\citep{chatgpt,openai2023gpt4,spark_AGI,DeepSeekR1}. 
Given proper instructions from humans, LLMs can generate desirable, high-quality text for multiple purposes. The great capability and versatility of LLMs have made the use and deployment of LLMs continue to expand, which contributes an exponentially growing proportion of the text data in the world.
OpenAI's ChatGPT reportedly produces approximately 100 billion words daily \cite{openai2023chatgpt}. In academia, an estimated 60,000 scholarly articles in 2023—over 1\% of all publications—were likely written with LLM assistance \cite{gray2023llm}, with 17.5\% of newly published computer science papers and 16.9\% of peer review texts incorporating LLM-generated content \cite{kannan2023research}. An analysis of over 6 billion sentences from Common Crawl further revealed that more than 57\% were machine-translated, highlighting the extent of automated content generation \cite{thompson2024shocking}. In fact, LLMs have also influenced various fields, including creative writing, healthcare, law, and software development. Authors use LLMs to co-author novels and generate narratives, as seen in Ayad Akhtar’s play ``McNeal'' \cite{akhtar2024mcneal}. In healthcare, they aid in diagnostics, automate documentation, and enhance medical education \cite{wang2023medical}. In the legal domain, LLMs assist in drafting legal documents and summarizing cases \cite{carrell2024legal}. Software development has also benefited from LLM-powered code generation, streamlining programming workflows \cite{chen2021codex}. 
In one of the largest dataset host platforms -- Huggingface,\footnote{\url{https://huggingface.co/}} we find that the LLM-generated data grows at an exponential speed, as shown in Fig.~\ref{fig:trend}. 
Furthermore, we fit a quadratic polynomial regression model, which predicts that the LLM-generated data will continue to expand up to 16,000 TB within the next decade. 
The exponential growth of LLM-generated text poses significant demands and challenges for efficiently storing LLM-generated data.


%
The rise of LLMs has introduced a new category of machine-generated data: \textit{LLM-generated text}.
Prior to the era of LLMs, machine-generated data usually referred to structured texts such as logs~\cite{ibm2013bigdata,bryant2008bigdata}.
For example, an IoT device can generate up to 1 TB per day in industrial applications, which are typically structured, containing numerical values and categorical information~\cite{xu2021iot}. 
As machine-generated text grows at a much higher speed than that from humans, a number of traditional compressors are designed to reduce the data storage volume of machine-generated text through the exploitation of the intrinsic structures~\cite{experimental2023zhang,time2022xiao, serve2022zhou}.
In contrast, LLM-generated text shares the trait of even more significantly rapid growth but exhibits distinct characteristics.

\begin{figure*}[t]
    \centering
    \includegraphics[width=0.9\linewidth]{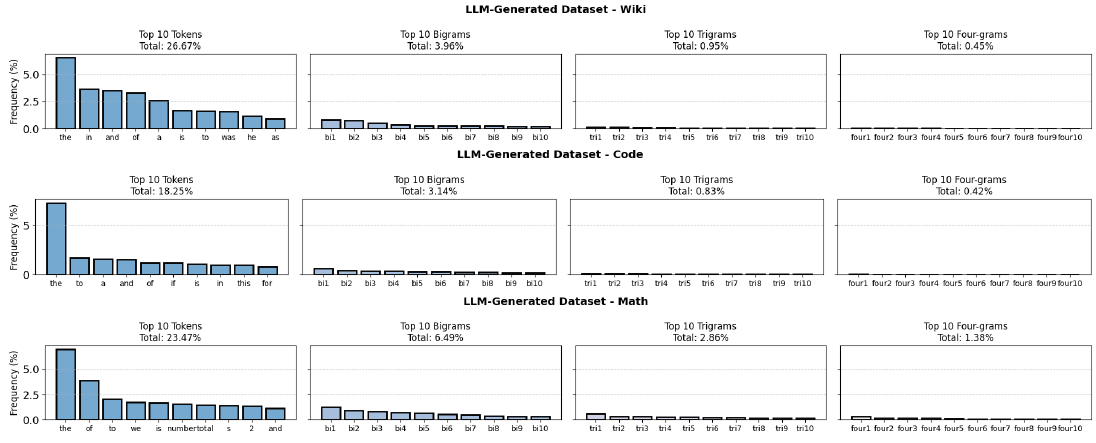} 
    \caption{N-gram frequency count for LLM-generated synthetic data.}
    \label{fig:ngram} 
\end{figure*}

Therefore, in this work, we conduct the first investigation of effective lossless compression approaches for LLM-generated data.
%
We begin with an in-depth analysis of the characteristics and compressibility of LLM-generated synthetic datasets. 
Specifically, we consider and curate 3 LLM-generated synthetic datasets from different domains. Based on those datasets, we measure the redundancy levels in LLM-generated text through n-gram statistics (Fig.~\ref{fig:ngram}), and find it is challenging to simply deduplicate the LLM-generated text.
Furthermore, we assess the compression potential by examining entropy per byte and mutual information (Table~\ref{tab:redundancy_analysis}),
which shows that LLM-generated data share similar levels of redundancy and information density to human-generated data.
Consequently, we find that traditional compression approaches, such as non-neural and neural compressors, can not achieve satisfactory compression ratios (Table~\ref{tab:compression_results}). Most methods achieve only a compression ratio of 3–6×, including entropy compression techniques based on neural networks.
Our findings provide an empirical foundation for designing efficient compression techniques \textit{tailored to the unique characteristics} of LLM-generated text.

To this end, we propose to incorporate LLMs to compress LLM-generated text. As LLMs are trained through next-token prediction, LLMs demonstrate high predictability on the LLM-generated data.
Hence, we can leverage the inherent predictability of LLMs to build powerful lossless compressors for LLM-generated data with remarkable compression capabilities. The need for \textit{compression rather than re-generation} arises from the randomness in LLM generation— even with the same prompt and model, exact re-generation remains infeasible. Therefore, an LLM can serve as a strong predictor for other LLM-generated data, but not with absolute accuracy. 
We conduct extensive experiments using 14 representative LLMs and 8 LLM-generated datasets spanning diverse categories.
The results demonstrate that LLM-based prediction methods achieve exceptional compression rates, exceeding 20×, which significantly surpasses the 3× compression ratio achieved by Gzip, the state-of-the-art lossless compression algorithm.
Furthermore, we show that the superior performance of LLM-based compressors generalizes across various LLM scales and dataset categories, highlighting their robustness and potential as a valuable tool for lossless data compression.
Finally, we also conduct an in-depth analysis of factors influencing compression efficiency, such as chunk size and data scale.

To summarize, this paper makes the following contributions:
\begin{itemize}
    \item To the best of our knowledge, we present the \textit{first systematic analysis} of the characteristics of LLM-generated data in terms of compressibility. As an emerging new category of machine-generated data, LLM-generated data exhibits significantly different characteristics from traditional machine-generated data while being similar to human-generated data. Consequently, traditional compressors can not effectively compress LLM-generated data. 
    \item We are also the first to propose leveraging LLMs to compress LLM-generated synthetic data. We show that LLMs have high inherent predictability for the LLM-generated data, inherited from the training with next-token prediction, despite differences in the training sources. It allows LLMs to compress the LLM-generated data effectively.  
    \item We adopt 8 datasets from different domains, and 14 state-of-the-art LLMs of different scales, trained on diverse data sources, and demonstrate that LLM-based compression consistently achieves remarkable compression rates. \textit{While conventional methods achieve only 3–5× compression, LLMs can reach nearly 23×}. We also conduct an in-depth analysis of the influence of various factors, such as different models, datasets, data sizes, and chunk sizes. Across various model sizes, datasets, data sizes, and chunk sizes, LLM-based compressors maintain significantly higher compression ratios than previous methods.
\end{itemize}
We hope our findings provide new insights and foundations for the management of text data in large-scale database.






\section{Background}

In this section, we begin by defining LLM- LLM-generated data and comparing it to conventional machine-generated data, emphasizing their core distinctions. We then examine the key characteristics of LLM-generated text in contrast to other datasets. We summarize the characteristics of different types of data in Table~\ref{tab:data_comparison}.
Finally, we delve into the role of randomness in LLM text generation, illustrating how stochastic elements within the model contribute to both diversity and uncertainty in the resulting outputs.
\begin{table}[ht]
    \centering
    \caption{Comparison of Machine-Generated Data, LLM-Generated Data, and Human-Generated Data.}
    \label{tab:data_comparison}
    \resizebox{\linewidth}{!}{%
    \begin{tabular}{lccc}
        \toprule
        \textbf{Characteristic}          & \textbf{Machine-Gen} & \textbf{LLM-Gen} & \textbf{Human-Gen} \\
        \midrule
        Automatically Generated          & \checkmark                     & \checkmark                             & --  \\
        Mimics Human Writing             & --                              & \checkmark                             & \checkmark  \\
        High Volume                      & \checkmark                     & \checkmark                             & --  \\
        Highly Structured                & \checkmark                     & --                                     & --  \\
        Time-Series Nature               & \checkmark                     & --                                     & --  \\
        Contains Semantic Context        & --                              & \checkmark                             & \checkmark  \\
        Factual \& Deterministic         & \checkmark                     & --                                     & --  \\
        \bottomrule
    \end{tabular}
    }
\end{table}
\subsection{Conventional Machine-Generated Data}

Machine-generated data refers to data that is entirely produced by machines or software without human intervention. The generation process is usually automated, occurring at fixed intervals or when triggered by specific events. The core characteristic of machine-generated data is that it originates from computational processes or hardware systems rather than being manually recorded. Examples include system logs, sensor readings, network traffic, and financial records. These data types are automatically generated by computational systems, IoT devices, and network infrastructure, often capturing real-time events and transactions without direct human input. They serve critical roles in monitoring, diagnostics, security analysis, and operational decision-making across industries~\cite{xu2021iot}.

\subsection{Large Language Model-Generated Data}

Synthetic data refers to data that is not derived directly from real-world events but is instead artificially generated by algorithms. LLM-generated synthetic data specifically refers to the text data created by Large Language Models (LLMs). Unlike traditional machine-generated data, which follows predefined rules, LLM-generated data mimics human language patterns, maintains contextual coherence, and exhibits linguistic diversity. Trained on massive corpora, LLMs generate text that closely resembles human writing, making them valuable for applications in content creation, data augmentation, and knowledge synthesis~\cite{gray2023llm,akhtar2024mcneal,wang2023medical,carrell2024legal,chen2021codex}.

Code-related datasets dominate with 571 datasets, aligning with the growing importance of code generation and LLMs in software development. Linguistics follows with 471 datasets, reflecting the widespread use of synthetic data in NLP tasks like text generation and translation. Other major categories include art (432 datasets), literature (151 datasets), and music (151 datasets), highlighting AI’s role in creative content generation. Science (246 datasets), math (165 datasets), and education (226 datasets) emphasize knowledge-based applications, while medical (131 datasets), business (38 datasets), finance (49 datasets), and law (77 datasets) demonstrate the increasing use of synthetic data in professional and regulatory fields.

\begin{figure}[htbp]
    \centering
    \includegraphics[width=0.95\linewidth]{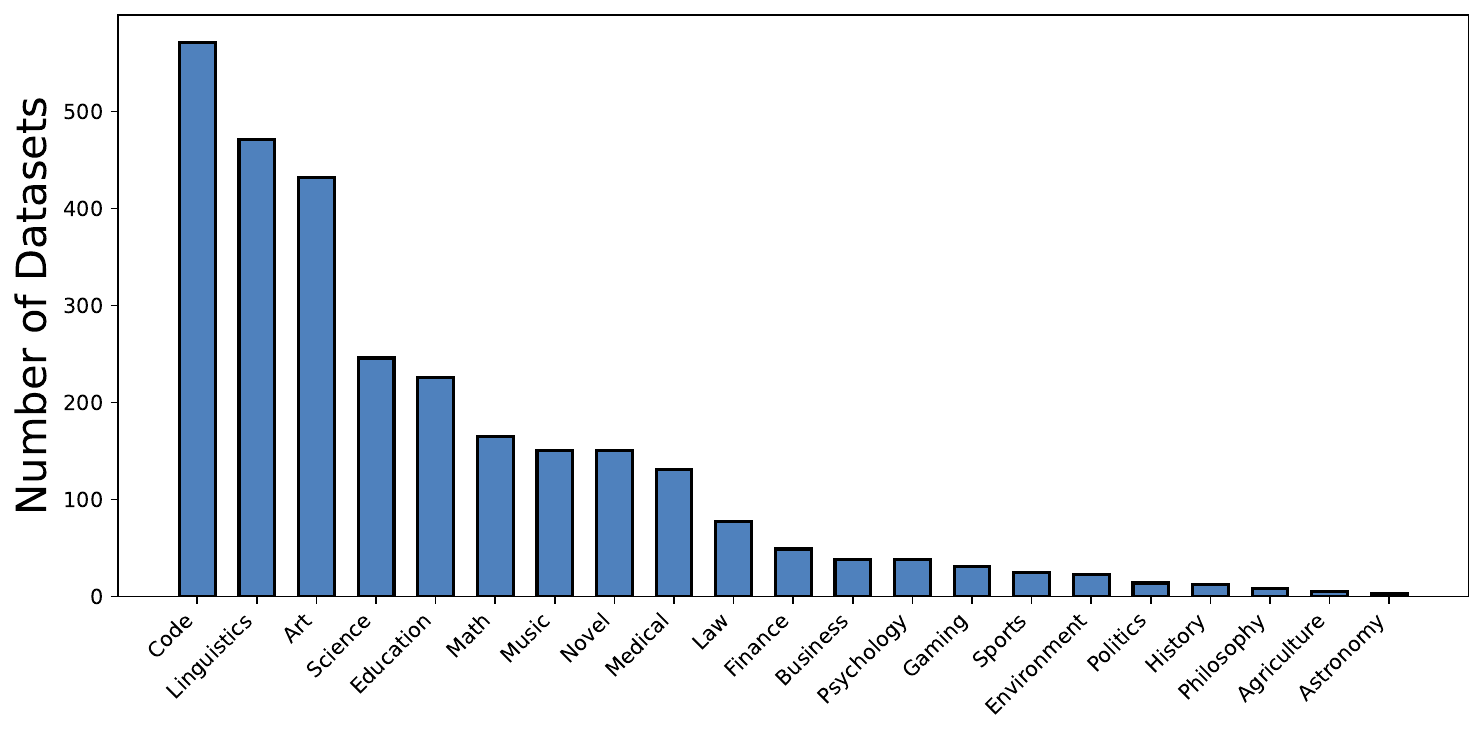} 
    \caption{LLM-generated data distribution.}
    \label{fig:synthetic_distribution} 
\end{figure}

\subsection{Characteristics of LLM-Generated Text vs. Other Datasets}
We begin by analyzing the characteristics of different types of data, aiming to gain insights that can guide more effective compression strategies. The characteristics of LLM-generated text differ significantly from traditional structured datasets used in database benchmarks. While machine-generated datasets such as TPC-H~\cite{TPC-H} are designed with well-defined schemas and structured query workloads, LLM-generated text exhibits high variability and redundancy and lacks explicit tabular organization. Similarly, human-generated text data—such as web articles, social media content, and historical corpora—shares some properties with LLM-generated text.

Table~\ref{tab:data_comparison} compares the key characteristics of machine-generated data, LLM-generated data, and human-generated data. Machine-generated data is highly structured, deterministic, and often time-series-based, whereas LLM-generated data mimics human writing and lacks the strict structure of traditional machine-generated formats. Unlike human-generated data, LLM-generated data is automatically produced and exhibits high volume but may not always be factually accurate.



\section{Compressibility Analysis of LLM-Generated Data}

In this subsection, we begin by exploring the potential for deduplication in LLM-generated data. We then evaluate the information content of machine-generated, human-generated, and LLM-generated data using entropy on different levels of tokenization and mutual information. Finally, we discuss the limitations of existing compression methods, setting the stage for new strategies that leverage the predictability of LLM-generated data.

\subsection{Deduplication Potential of LLM-Generated Data}
Deduplication is a common technique for machine-generated data, such as tabular or time-series data~\cite{time2022xiao,experimental2023zhang}. To assess its applicability to LLM-generated data, we performed an n-gram analysis on LLM-generated synthetic datasets to evaluate their redundancy and deduplication potential.
By examining the frequency distribution of tokens, bigrams, trigrams, and four-grams across different domains, we aim to obtain a holistic view of the redundancy of these datasets, i.e., how much repeated content exists in these datasets and whether deduplication could effectively reduce storage or improve compression efficiency.

As illustrated in Fig.~\ref{fig:ngram}, the results reveal a low level of redundancy across all datasets. In the clinical dataset, for example, the top 10 tokens account for 27.53\% of the text, but bigrams, trigrams, and four-grams contribute significantly less, with bigrams at 5.03\%, trigrams at 1.86\%, and four-grams at just 0.93\%. A similar pattern is observed in the code and math datasets, where repeated n-grams take up only a small portion of the text.

These findings suggest that deduplication techniques, which have good performance on machine-generated data, have limited potential for LLM-generated synthetic datasets.
Unlike conventional machine-generated data, which often contains repetitive phrases and structural redundancies, LLM-generated text is more diverse, reducing the effectiveness of deduplication-based compression methods. 
One possible scenario in the future is that when the same LLM generates text from different sources, there may be opportunities for file-level deduplication. However, this falls beyond the scope of the research topic discussed in this paper.
%

\subsection{Information Evaluation of the machine-generated data, human-generated data, and LLM-generated data.}

Lossless compression aims to encode a sequence of symbols \( x_{1:n} \) into a bitstream with minimal expected length while ensuring perfect reconstruction of the original data. According to Shannon's source coding theorem~\cite{shannon1948mathematical}, the optimal compression efficiency is bounded by the entropy of the source distribution. Given a source probability distribution \( \rho(x) \), the Shannon entropy is defined as:
\begin{equation}
H(\rho) := \mathbb{E}_{x \sim \rho}[-\log_2 \rho(x)]
\end{equation}

The Shannon entropy represents the theoretical lower bound on the average number of bits needed to encode a symbol drawn from \( \rho \). A well-designed compression scheme assigns shorter codes to more frequent symbols and longer codes to rare symbols, achieving an average code length \( L \) satisfying:
\begin{equation}
L \geq H(\rho)
\end{equation}
where equality holds under an optimal entropy coder such as Huffman coding or Arithmetic coding in the limit of large sequences. This entropy bound establishes the fundamental limits of lossless data compression, beyond which further compression requires exploiting higher-order dependencies or introducing loss.

To evaluate the impact of tokenization granularity on information representation, we compute the entropy per byte for character-level, subword (BPE), and word-level tokenization. 
The entropy per byte is computed as follows:
\[
H_{\text{byte}} = \frac{H_{\text{token}}}{L_{\text{avg}}}
\]
where\( H_{\text{token}} \) is the Shannon entropy per token, calculated as:
  \[
  H_{\text{token}} = - \sum_{t \in T} P(t) \log_2 P(t)
  \]
  where \( T \) is the set of tokens, and \( P(t) \) is the probability of token \( t \). \( L_{\text{avg}} \) is the average token length in bytes, defined as:
  \[
  L_{\text{avg}} = \frac{\sum_{t \in T} P(t) \cdot L(t)}{\sum_{t \in T} P(t)}
  \]
  where \( L(t) \) is the byte length of token \( t \). This formula applies to character-level, subword (BPE), and word-level tokenization by adjusting the token set \( T \) accordingly.

We also use \textit{Mutual Information Between Consecutive Words}, which measures how much knowing one word helps predict the next word. It is computed as:
\[
MI(W_i, W_{i+1}) = \sum_{w_i \in W} \sum_{w_{i+1} \in W} P(w_i, w_{i+1}) \log_2 \frac{P(w_i, w_{i+1})}{P(w_i) P(w_{i+1})}
\]
where \( W_i \) and \( W_{i+1} \) are consecutive words in the dataset, \( P(w_i, w_{i+1}) \) is the joint probability of two consecutive words \( w_i \) and \( w_{i+1} \), and \( P(w_i) \) and \( P(w_{i+1}) \) are the marginal probabilities of words \( w_i \) and \( w_{i+1} \), respectively. This metric provides insights into the structural dependencies within different tokenization schemes, helping assess how well each method retains contextual information.



Table~\ref{tab:redundancy_analysis} presents a comparative analysis of these characteristics, where Char-E, BP-E, and W-E indicate Char-Entropy, BPE-Entropy and Word-Entropy. To ensure a fair comparison, the human-generated dataset and the LLM-generated dataset both originate from the same source (Wikipedia), with each sentence conveying the same meaning. The only difference is that the LLM-generated data has been rewritten by an LLM. For the machine-generated data, i.e., TPC-H, we collected the comments field from its tables to ensure a fair comparison with the other two datasets.

\begin{table}[h]
    \centering
    \caption{Redundancy and token frequency characteristics of LLM-generated text vs. structured datasets.}
    \begin{tabular}{c|cccc}
        \hline
        \textbf{Dataset} & \textbf{Char-E} & \textbf{BP-E} & \textbf{W-E} & \textbf{Mutual Info} \\
        \hline
        LLM-Generated & 4.67 & 2.51 & 2.57 & 2.95\\
        Human-Generated & 4.63 &  2.53 & 2.59 & 2.73\\
        TPC-H & 4.34 & 1.72 & 0.94 & 1.23\\
        \hline
    \end{tabular}
    \label{tab:redundancy_analysis}
\end{table}

We observe that the entropy measures of both LLM-generated data and human-generated data are similar across character, BPE, and word levels. Moreover, these entropy values are relatively low, indicating that traditional compression methods, including state-of-the-art approaches like Zstd, are likely to achieve only relatively limited compression ratios.

Interestingly, when examining mutual information (Mutual Info), LLM-generated data exhibits relatively higher token predictability, despite each sentence conveying the same meaning as its human-generated counterpart. As expected, TPC-H shows very low mutual information since it is randomly generated and lacks a continuous semantic structure.

\subsection{Limitations of Existing Compression Methods}
Built upon the aforementioned analysis, we selected three types of LLM-generated synthetic text for a preliminary examination of traditional compression methods, including a mixed-domain Wikipedia dataset and two specific-domain datasets: math and code. For compression baselines, we used three state-of-the-art compressors—LZMA~\cite{lz77}, Gzip~\cite{gzip}, and Zstd~\cite{zstd}, along with three Neural Network-based (NN-based) compressors—NNCP~\cite{bellard2023nncp}, TRACE~\cite{mao2022trace}, and PAC~\cite{mao2023faster} to conduct a comprehensive study on the compressibility of LLM-generated text.

\begin{table}[h]
    \centering
    \caption{Compression ratios of different datasets using Gzip, LZMA, Zstd-22, NNCP, TRACE and PAC.}
    \begin{tabular}{c|cccccc}
        \hline
        \textbf{Dataset} & \textbf{Gzip} & \textbf{LZMA} & \textbf{Zstd} & \textbf{NNCP} & \textbf{TRACE} & \textbf{PAC}\\
        \hline
        Wiki   & 2.73 & 4.24 & 4.17 & 5.04 & 4.11 & 4.17 \\
        Code  & 3.47 & 4.90 & 4.85 & 6.30 & 4.53 & 5.08 \\
        Math  & 4.30 & 5.99 & 5.90 & 7.76 & 5.34 & 6.39 \\
        \hline
    \end{tabular}
    \label{tab:compression_results}
\end{table}

The compression results are given in Table~\ref{tab:compression_results}. We can find that, while modern compressors such as LZMA and neural-based methods like NNCP and PAC outperform traditional dictionary-based approaches such as Gzip and Zstd-22, the overall compression ratios remain relatively limited. Among the traditional methods, LZMA achieves the highest compression efficiency, with ratios ranging from 4.24 (Wiki) to 5.99 (Math). Zstd-22 performs similarly to LZMA but offers slightly lower compression ratios across all datasets. Neural-based methods demonstrate further improvements, with NNCP consistently achieving the best performance among the listed approaches, obtaining 6.30 (Code), 7.76 (Math), and 5.04 (Wiki). TRACE and PAC provide competitive results but still fall within a similar compression range.  

Despite these incremental improvements, all existing methods struggle to achieve compression ratios beyond a single-digit range. The results highlight the inherent limitations of conventional entropy and dictionary-based compressors, as well as the current generation of neural compressors, when applied to unstructured and irregular text datasets. It calls for developing efficient compression techniques tailored to the unique characteristics of the LLM-generated text.

\section{LLM-based Compression Framework}
To resolve the irregularity of the LLM-generated text, we propose to leverage LLMs to compress the LLM-generated text. In this section, we present a detailed LLM-based framework that achieves remarkable lossless compression performance compared to traditional compression approaches.

\subsection{Conceptual Framework}
In this section, we provide a conceptual description of our LLM-based compression framework. It consists of two steps: a) context-based prediction and b) encoding coding.

\begin{figure*}[htbp]
    \centering
    \includegraphics[width=0.85\linewidth]{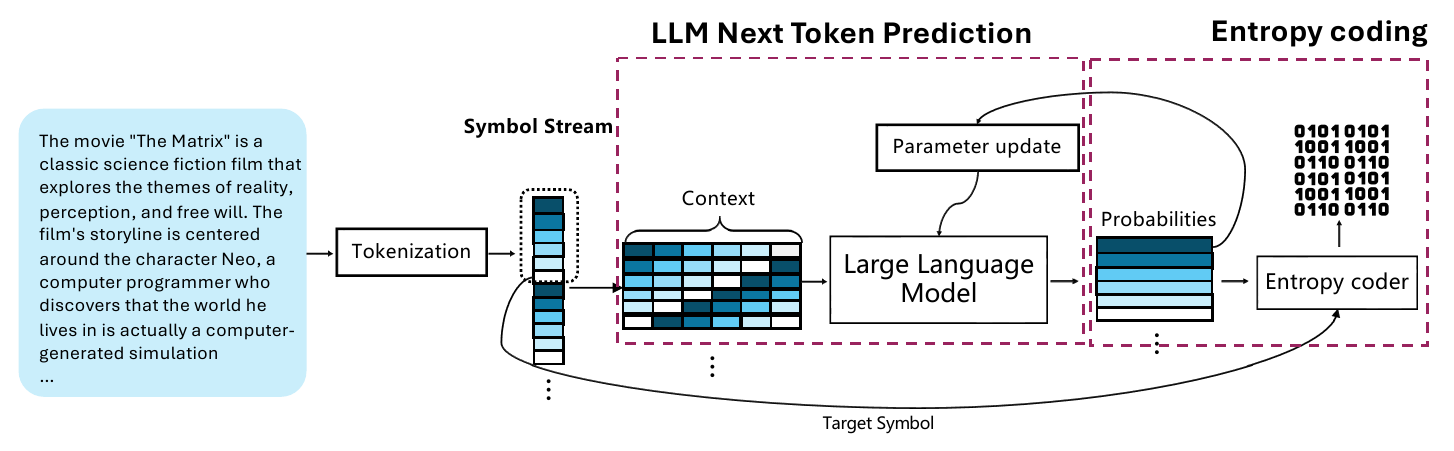} 
    \caption{General process of Neural-based Compression.}
    \label{fig:example} 
\end{figure*}

\subsubsection{Context-based prediction} Context-based prediction is a foundational technique in neural lossless compression\cite{bellard2023nncp, Deletang2024, mao2022trace}, aiming to estimate the probability distribution of a symbol \( x_t \) conditioned on its preceding context \( x_{<t} \). Accurate probability estimation affords efficient entropy coding and reduces the redundancy in data representation. 

Given a sequence of symbols \( X = (x_1, x_2, \dots, x_T) \) in length $T$, the probability distribution of the entire sequence can be factorized using the chain rule of probability:
\begin{equation}
P(X) = \prod_{t=1}^{T} P(x_t \mid x_{<t}),
\end{equation}
where \( x_{<t} = (x_1, x_2, \dots, x_{t-1}) \) represents the preceding symbols. The objective of context-based prediction is to accurately model the conditional probability \( P(x_t \mid x_{<t}) \) to minimize the entropy of the encoded sequence.

\subsubsection{Entropy coding} Entropy coding relies on the fundamental principle that the optimal encoding length of a symbol \( x_t \) is given by its self-information:
\begin{equation}
L(x_t) = -\log_2 P(x_t \mid x_{<t}).
\end{equation}

Thus, the expected length per symbol, also known as the entropy, is given by:

\begin{equation}
\begin{aligned}
    H(X) &= \mathbb{E}[-\log_2 P(x_t \mid x_{<t})] \\
    &= - \sum_{x_t} P(x_t \mid x_{<t}) \log_2 P(x_t \mid x_{<t}).
\end{aligned}
\end{equation}

Efficient entropy coding is achieved when the estimated probability distribution \( \hat{P}(x_t \mid x_{<t}) \) closely matches the true probability distribution \( P(x_t \mid x_{<t}) \), leading to minimal redundancy in encoding. On the other hand, the coding inefficiency \( R \) can be quantified as:
\begin{equation}
R = H(\hat{P}) - H(P),
\end{equation}
where \( H(\hat{P}) \) is the entropy computed using the estimated probability distribution. The Kullback-Leibler (KL) divergence provides a measure of the discrepancy between these distributions:
\begin{equation}
D_{KL}(P \parallel \hat{P}) = \sum_{x_t} P(x_t \mid x_{<t}) \log_2 \frac{P(x_t \mid x_{<t})}{\hat{P}(x_t \mid x_{<t})}.
\end{equation}

From information theory, it follows that:
\begin{equation}
H(P) \leq H(\hat{P}) = H(P) + D_{KL}(P \parallel \hat{P}).
\end{equation}

This inequality implies that a more accurate probability model reduces \( D_{KL} \), leading to a smaller gap between the estimated entropy \( H(\hat{P}) \) and the true entropy \( H(P) \), thereby improving compression efficiency. In the ideal case where \( \hat{P} = P \), the KL divergence becomes zero, achieving optimal entropy coding.








\subsection{The Next-Token Prediction Framework}

LLMs generate text in an autoregressive manner. In other words, LLMs predict the next token in a sequence based on previously generated tokens.
The training of LLMs leverages the next-token prediction objective, which is to model the probability distributions over sequences of tokens. 

The training process of LLMs essentially resembles the context-based prediction. 
To capture relevant contextual information, LLMs employ an embedding function \( f_{\theta} \), parameterized by \( \theta \), to transform the context \( x_{<t} \) into a hidden state representation \( h_t \):
\begin{equation}
 h_t = f_{\theta}(x_{<t}).
\end{equation}

Transformer-based Models (e.g., GPT-like Architectures) utilize self-attention mechanisms to capture long-range dependencies~\cite{transformers}:
\begin{equation}
  h_t = \sum_{j < t} \alpha_{t,j} (W_V x_j),
\end{equation}
where attention weights \( \alpha_{t,j} \) are computed as follows:
\begin{equation}
  \alpha_{t,j} = \frac{\exp((W_Q x_t)^\top (W_K x_j))}{\sum_{j < t} \exp((W_Q x_t)^\top (W_K x_j))}.
\end{equation}
  
Once the contextual representation \( h_t \) is obtained, the probability distribution of the next token \( x_t \) is modeled using a softmax function:
\begin{equation}
P(x_t \mid x_{<t}) = \text{softmax}(W_o h_t),
\end{equation}
where \( W_o \) is a learnable weight matrix that maps the hidden state to output logits. The predicted probability distribution is then utilized in entropy coding algorithms to achieve optimal lossless compression. In this work, we implement the entropy coding as the arithmetic coding~\cite{arithmetic}.

\paragraph{Tokenization and Embedding.} Prior to processing, text is tokenized into discrete units using methods such as Byte Pair Encoding (BPE) or WordPiece. Each token $x_t$ is then mapped to a high-dimensional embedding vector $e_t$ via a pre-trained embedding matrix:
\begin{equation}
e_t = \text{Embedding}(x_t).
\end{equation}

\paragraph{Transformer-Based Processing.} The token embeddings are processed through multiple layers of a Transformer model, which consists of self-attention and feedforward networks. The self-attention mechanism computes token dependencies using:
\begin{equation}
\text{Attention}(Q, K, V) = \text{softmax} \left( \frac{QK^T}{\sqrt{d_k}} + M \right) V,
\end{equation}
where $Q$, $K$, and $V$ are query, key, and value matrices, respectively, and $M$ is a causal mask preventing future tokens from influencing current token computations.

\paragraph{Probability Estimation and Decoding.} The final hidden representation $h_t$ of the last Transformer layer is projected onto the vocabulary space and converted into a probability distribution using a softmax function:
\begin{equation}
P(x_{t+1} | x_1, \dots, x_t) = \text{softmax}(W h_t + b),
\end{equation}
where $W$ and $b$ are trainable parameters.

\paragraph{Training Objective}
The model is trained using a causal language modeling objective, or next-token prediction, minimizing the cross-entropy loss:
\begin{equation}
L = -\sum_{t=1}^{T} \log P(x_t | x_1, \dots, x_{t-1}).
\end{equation}

Gradients are computed via backpropagation, and model parameters are updated using optimization algorithms such as Adam~\citep{adam}.

\subsection{Entropy Coding}
We use arithmetic coding as the entropy coder~\cite{arithmetic}. Arithmetic coding is an advanced entropy encoding technique that efficiently represents a sequence of symbols as a single floating-point number within the unit interval \([0,1)\). Unlike Huffman coding, which assigns fixed-length binary codes to individual symbols~\cite{huffman1952method}, arithmetic coding progressively refines an interval based on symbol probabilities, achieving near-optimal compression close to the Shannon entropy limit.

\subsubsection{Encoding Process}
Given an input sequence \( X = (x_1, x_2, \dots, x_T) \) in length $T$ with respective probabilities \( P(x_t) \), arithmetic coding proceeds as follows. Initially, the encoding interval is set to \([L, H) = [0,1)\). 

For each symbol \( x_t \) in the sequence, the current interval width is computed as:
\[
R = H - L.
\]
The interval is then updated based on the cumulative probability of preceding symbols:
\[
L' = L + R \cdot F(x_t), \quad H' = L + R \cdot (F(x_t) + P(x_t)),
\]
where \( F(x_t) \) represents the cumulative probability of all symbols preceding \( x_t \). 

This process continues iteratively until all symbols have been encoded. The final representation is chosen as a value within the interval \([L_T, H_T)\), ensuring a minimal binary representation.

\subsubsection{Decoding Process}
Decoding follows the inverse procedure. The interval is initialized as \([L, H) = [0,1)\), and the encoded value \( X \) is read. 

At each step, the symbol \( x_t \) is identified as the one whose assigned sub-interval \([L', H')\) contains \( X \). The decoded symbol is then output, and the interval is updated using the same transformation as in the encoding process. 

To ensure correct symbol retrieval in subsequent steps, \( X \) is normalized within the updated interval, and the process continues until the entire sequence is reconstructed.


\subsection{Compression or Generation?}
One might be curious about the necessity of using LLMs to compress LLM-generated instead of directly regenerating the text.
In most cases, LLMs do not generate identical outputs for the same input due to multiple interacting factors that introduce variability. While LLMs are fundamentally deterministic in that they map a given input sequence to a set of probability distributions over tokens, the presence of stochastic sampling, architectural routing mechanisms, and computational nuances can lead to nondeterministic behavior. Even when all explicit sources of randomness, such as temperature and sampling, are eliminated, underlying numerical precision issues and architectural decisions may still cause slight variations in the generated text.

\paragraph{Temperature.}
One of the primary factors contributing to output variability is the temperature ($\tau$), a parameter that modulates the probability distribution over possible next tokens. Mathematically, temperature scales the logits ($z_i$) corresponded to word ($w_i$), before applying the softmax function:
\begin{equation}
P(w_i) = \frac{e^{z_i/\tau}}{\sum_j e^{z_j/\tau}}.
\end{equation}
When $\tau > 1$, the probability distribution becomes flatter, increasing the likelihood of selecting lower-probability tokens, which enhances diversity in the generated output. Conversely, as $\tau \to 0$, the softmax function increasingly favors the token with the highest probability, making the output more deterministic \citep{holtzman2020curious}. However, due to floating-point precision limitations, even a near-zero temperature may not always yield perfectly consistent results. 

\paragraph{Sampling.}
In addition to temperature, sampling strategies also significantly influence output determinism. Techniques such as top-k sampling restrict token selection to the $k$ most probable candidates \citep{fan2018hierarchical}, while top-p (nucleus) sampling dynamically adjusts the candidate set to include the smallest subset of tokens whose cumulative probability exceeds a threshold $p$ \citep{holtzman2020curious}. Additionally, even deterministic decoding methods such as greedy decoding (where the model selects the most probable token at each step) or beam search\citep{freitag2017beam} (which maintains multiple hypotheses before selecting the highest-probability sequence) can exhibit variations due to numerical rounding errors, computational parallelism, and subtle differences in implementation.

\paragraph{Architecture.}
The model’s architecture can also cause the issue of non-determinism. In traditional dense Transformer models, such as GPT-3~\citep{gpt3} and LLaMA~\citep{touvron2023llama}, every token undergoes a fixed computational pathway, ensuring a higher degree of determinism when sampling is disabled \citep{touvron2023llama}. However, more recent models, such as GPT-4~\citep{openai2023gpt4}, are believed to employ Mixture-of-Experts (MoE) architectures, where only a subset of specialized neural network components, or ``experts'' are activated for each token~\citep{adaptive_moe}. The routing mechanism that determines which experts are used can introduce variability for its probabilistic nature, which can be sensitive to minute numerical fluctuations \citep{lepikhin2021gshard}.


\begin{table*}[t]
\centering
\caption{Detailed Characteristics of Selected Large Language Models}
\label{tab:llm_comparison}
\begin{adjustbox}{max width=\textwidth}
\begin{tabular}{lccccccc}
    \toprule
    \textbf{Model} & \textbf{Size (B)} & \textbf{Type} & \textbf{Year} & \textbf{Organization} & \textbf{Context Length} & \textbf{Data Source} & \textbf{Use Case} \\
    \midrule
    \texttt{OpenELM-1\_1B}~\cite{mehtaOpenELMEfficientLanguage2024}  & 1.1B  & Decoder & 2024 & Apple    & 4096   & Web, Books & Efficient NLP tasks \\
    \texttt{AMD-OLMo-1B}~\cite{AMD-OLMo}    & 1B    & Decoder & 2024 & AMD      & 4096   & Code, Web Data & Optimized AI applications \\
    \texttt{Llama-3.2-1B}~\cite{llama3.2}   & 1B    & Decoder & 2024 & Meta AI  & 8192   & Web, Academic Papers & General NLP \\
    \texttt{Llama-3.2-1B-Instruct}~\cite{llama3.2} & 1B  & Decoder & 2024 & Meta AI  & 8192   & Web, Academic Papers & Instruction-tuned NLP \\

    \texttt{Llama-3.2-3B}~\cite{llama3.2}   & 3B    & Decoder & 2024 & Meta AI  & 8192   & Web, Books, Papers & Open-source NLP \\
    \texttt{Llama-3.2-3B-Instruct}~\cite{llama3.2} & 3B  & Decoder & 2024 & Meta AI  & 8192   & Web, Books, Papers & Instruction-tuned NLP \\
    \texttt{Llama-3.1-8B}~\cite{llama3.1}   & 8B    & Decoder & 2024 & Meta AI  & 16384  & Large-scale Corpora & Advanced reasoning \\
    \texttt{Llama-3.1-8B-Instruct}~\cite{llama3.1} & 8B  & Decoder & 2024 & Meta AI  & 16384  & Large-scale Corpora & Instruction-following NLP \\
    \texttt{Qwen2.5-14B}~\cite{qwen2.5}    & 14B   & Decoder & 2024 & Alibaba  & 32768  & Multilingual Corpora & General NLP \\
    \texttt{Qwen2.5-14B-Instruct-1M}~\cite{qwen2.5} & 14B & Decoder & 2024 & Alibaba  & 32768  & Multilingual Corpora & Instruction-tuned NLP \\\hdashline[1pt/1pt]
    \texttt{Qwen2.5-Math-1.5B}~\cite{yang2024qwen25mathtechnicalreportmathematical} & 1.5B & Decoder & 2024 & Alibaba & 16384 & Mathematical Data & Mathematical reasoning \\
    \texttt{Rho-Math-1B}~\cite{lin2024rho1} & 1B & Decoder & 2024 & Microsoft & 2048 & Mathematical Data & Mathematical problem \\
    \texttt{Qwen2.5-Coder-1.5B}~\cite{hui2024qwen25coder} & 1.5B & Decoder & 2024 & Alibaba & 16384 & Code Corpora & Code generation \\
    \texttt{DeepDeek-Coder-1.3b-Instruct}~\cite{deepseek-coder} & 1.3B & Decoder & 2024 & DeepSeek AI & 16000 & Code, Natural Language & Code completion \\
    \bottomrule
\end{tabular}
\end{adjustbox}
\end{table*}

\section{Evaluation}

In this section, we systematically evaluate the compression performance of various methods across multiple LLM-generated datasets. We begin by outlining the experimental setup. Next, we introduce the baseline methods, including traditional entropy-based, dictionary-based, and NN-based compression techniques, providing a comprehensive comparison against our proposed approach.  

To better understand key factors influencing compression efficiency, we conduct in-depth analyses on chunk size, model size, and dataset scale, investigating their impact on compression ratios. Additionally, we compare LLMs with and without instruction tuning, exploring how post-training affects LLM-based compression. Finally, we categorize the data based on structural properties and statistical characteristics, offering insights into how different types of text influence compression efficiency.

\subsection{Experimental Setups}
In this section, we detail the datasets, preprocessing steps, and evaluation metrics used in our study. 

\subsubsection{Evaluation Datasets}

\paragraph{Wiki~\cite{sivesind_2023}.}This dataset consists of human-written and LLM-generated Wikipedia introductions. The human-written samples are sourced from Wikipedia, while the machine-generated samples are produced using the GPT-3 model.

\paragraph{Article~\cite{sivesind_2023}.}The Article dataset includes scientific research abstracts in both human-written and machine-generated formats. Human-written abstracts are sourced from existing scientific publications, while LLM-generated abstracts are produced using GPT-3.5. 

\paragraph{Code~\cite{synthetic_code_generations}.}This dataset was generated using Mixtral-8x7B. It contains programming problems and solutions across multiple languages, including Python, JavaScript, TypeScript, C++, and C. 

\paragraph{Math~\cite{mitra2024orcamath}.}The Math dataset contains approximately 200K grade school math word problems, designed to improve the mathematical reasoning capabilities of language models. The answers were generated using Azure GPT-4-Turbo, ensuring consistency across different types of problems.

\paragraph{Science~\cite{li2023camel}.}The Science dataset consists of 20K problem-solution pairs across 25 physics topics, with 25 subtopics per topic and 32 problems per topic-subtopic pair. All problem-solution pairs were generated using GPT-4, making this dataset well-suited for evaluating the compression of structured scientific content.

\paragraph{Clinical~\cite{kweon2023publicly}.}This dataset is the official dataset for Asclepius (\texttt{arXiv}), formatted in a Clinical Note - Question - Answer structure. Synthetic clinical notes were generated from PMC-Patient case reports using GPT-3.5, followed by instruction-answer pairs for 157K synthetic discharge summaries. 

\paragraph{Web.}The Web dataset consists of movie critic reviews generated using ChatGPT~\cite{openai2023chatgpt}, structured to mimic human-written reviews.

\paragraph{Novel.}The Novel dataset consists of a travel book generated using LongWriter~\cite{bai2024longwriter}, simulating the style and structure of human-authored long-form narratives.

\subsubsection{Evaluation Metrics}
For lossless compression, the compression ratio is the primary performance metric. It quantifies the reduction in data size achieved by compression and is defined as the ratio between the original data size and the compressed data size:
\begin{equation}
R = \frac{S_{\text{original}}}{S_{\text{compressed}}}
\end{equation}
where \( S_{\text{original}} \) is the size of the uncompressed data, and \( S_{\text{compressed}} \) is the size of the data after applying a compression algorithm. A higher compression ratio indicates greater space savings, making storage and transmission more efficient.

\subsubsection{Hardware Platform}
Our experiment is conducted on a cluster powered by an AMD EPYC 7742 64-Core Processor equipped with a NVIDIA A100 Tensor Core GPU.

\subsection{Baselines}

To thoroughly investigate the compression characteristics of LLM-generated data, we introduce nine baselines spanning three categories: entropy-based, dictionary-based, and neural-based methods. Additionally, to examine the impact of different LLMs on the proposed framework, we include 14 LLMs in this comprehensive study.

\subsubsection{Entropy-based Compressor}

Huffman coding, Arithmetic coding and Finite State Entropy (FSE) is selected as entropy-bassed baselines.
Huffman coding~\cite{huffman1952method} is a foundational entropy coding technique, assigns shorter codes to more frequent symbols and serves as the basis for many compression algorithms. 
Finite State Entropy (FSE)~\cite{colletFSE} is an entropy coding method that dynamically models symbol probabilities using a finite-state machine, improving efficiency over fixed-length coding schemes like Huffman coding.
Arithmetic coding~\cite{arithmetic} represents an entire message as a fractional number in the interval $[0,1)$, refining the interval based on symbol probabilities to achieve near-optimal compression.

\subsubsection{Dictionary-based Compressor}

LZMA (Lempel-Ziv-Markov chain algorithm)~\cite{lz77} extends LZ77 with probability modeling and range coding, achieving high compression ratios at the cost of increased computational complexity. Gzip~\cite{gzip}, based on the DEFLATE algorithm, integrates LZ77 for dictionary compression with Huffman coding for entropy encoding, offering a well-balanced tradeoff between compression ratio and speed, widely used for file compression and web data transfer. Zstandard (Zstd)~\cite{zstd}, developed by Facebook, is a high-performance compression algorithm that combines LZ77-style compression with Finite State Entropy (FSE) coding, balancing speed and compression ratio while supporting dictionary-based compression. 

\subsubsection{NN-based Compressor}

We adopt NNCP~\cite{bellard2023nncp}, TRACE~\cite{mao2022trace} and PAC~\cite{mao2023faster} as our NN baselines. 
Neural Network-based Compression (NNCP) utilizes deep learning models, such as autoencoders and transformers, to learn compact representations of data.
TRACE designed a slim transformer to accelerate the compression procedure. PAC further builds an order model to replace attention to achieve faster compression speed.
\begin{table*}[t]
\vspace{-0.15in}
\belowrulesep=0pt
\aboverulesep=0pt
	\centering
\caption{Compression results for various LLM-generated synthetic datasets.}
\label{tab:compression_results_datasets}
	\begin{tabular}{l|*{8}{c}}
		\toprule
		\rule{0pt}{10pt}\multirow{3}{*}{Method}&\multicolumn{8}{|c}{Entropy-based Compressor} \\
         \cmidrule{2-9}
		 & Wiki & Code & Math & Clinical & Web & Science & Novel & Artical  \\
		\midrule
		Huffman &$1.69$ &	$1.62$&	$1.65$&	$1.82$&	$1.77$&	$1.64$&	$1.76$&	$1.81$	\\
        Arithmetic     &$1.70$ & $1.63$ & $1.66$ & $1.61$ & $1.77$ &$1.64$ &$1.77$ & $1.81$
	\\
        FSE    &$1.68$ &	$1.62$&	$1.65$&	$1.83$&	$1.77$&	$1.64$&	$1.75$&	$1.80$\\
        \midrule

        \rule{0pt}{10pt}\multirow{3}{*}{Method}&\multicolumn{8}{|c}{Dictionary-based Compressor} \\
         \cmidrule{2-9}
		 & Wiki & Code & Math & Clinical & Web & Science & Novel & Artical  \\
		\midrule
        Gzip    &$2.73$&	$3.47$&	$4.30$&	$5.70$&	$4.26$&	$5.20$&	$3.40$&	$3.48$	\\
        Lzma    &$4.24$&	$4.90$&	$5.99$&	$9.76$&	$6.49$&	$7.13$&	$5.74$&	$5.34$	\\
        Zstd-22  &$4.17$&	$4.85$&	$5.90$&	$9.95$&	$6.46$&	$7.05$&	$5.68$&	$5.31$\\
        \midrule

        \rule{0pt}{10pt}\multirow{3}{*}{Method}&\multicolumn{8}{|c}{Neural-based Compressor} \\
         \cmidrule{2-9}
		 & Wiki & Code & Math & Clinical & Web & Science & Novel & Artical  \\
		\midrule
	NNCP &$5.04$	&$6.30$	&$7.76$	&$12.56$	&$2.95$	&$7.38$	&$5.39$	&$6.30$	\\
        TRACE & $4.11$ & $4.53$ & $5.34$ & $9.55$ & $4.75$ & $5.09$ & $5.09$ & $5.53$
	\\
        PAC    & $4.17$ & $5.08$ & $6.39$ & $9.94$ & $3.48$ & $6.66$ & $5.02$  &$5.63$\\
        Ours  & $\mathbf{14.62}$ & $\mathbf{20.44}$ & $\mathbf{15.22}$ & $\mathbf{17.84}$ & $\mathbf{22.26}$ & $\mathbf{23.80}$ & $\mathbf{23.13}$ & $\mathbf{16.97}$\\
        \midrule
            
        \bottomrule
	\end{tabular}
\end{table*}

\subsubsection{LLM-based Compressor.}

In this paper, we incorporate LLMs of different sizes and from various companies, covering a range from 1B to 14B parameters.  
In deep learning, 250M, 1B, and 11B refer to models with 250 million, 1 billion, and 11 billion parameters, respectively. Each parameter is typically stored as a 16-bit (FP16) or 32-bit (FP32) floating-point number.


The selected models provide a diverse comparison across different model sizes and organizations. Specifically, we included multiple 1B-sized models, OpenELM-1.1B, AMD-OLMo-1B, and Llama-3.2-1B, to evaluate performance under similar computational constraints. The Llama series was chosen to investigate scaling laws in compression tasks, given its variations in size and training methodology. Furthermore, models from various companies, including Apple, AMD, Meta AI, and Alibaba, were selected to assess the generalization of our proposed method across different architectures and datasets. 

Additionally, we include instruction-tuned versions of several models, which are aligned to human preferences, including Llama-3.2-1B-Instruct, Llama-3.2-3B-Instruct, Llama-3.1-8B-Instruct, and Qwen2.5-14B-Instruct-1M. Compared to their base counterparts, these models demonstrate enhanced usability in real-world interactive AI applications. In the following experiments, unless otherwise specified, we use Llama-3.1-8B as the base LLM.

\subsection{Overview of Compression Performance Across LLM-Generated Datasets}

In this subsection, we provide a comprehensive analysis of compression efficiency across various LLM-generated synthetic datasets. Table~\ref{tab:compression_results_datasets} presents the results obtained using different categories of compression methods, including entropy-based, dictionary-based, and neural-based approaches. The evaluation covers a diverse range of datasets, including Wiki, Code, Math, Clinical, Web, Science, Novel, and Article, allowing for a thorough comparison of how different methods handle distinct types of LLM-generated text.  

The entropy-based compressors, including Huffman coding, Arithmetic coding, and FSE coding, exhibit the lowest compression performance across all datasets. Their limited ability to exploit long-range dependencies in text restricts their effectiveness, resulting in compression ratios consistently below 2.0. This highlights the fundamental limitation of statistical coding techniques when applied to highly structured, synthetic text.  

Dictionary-based methods, such as Gzip, LZMA, and Zstd-22, show a significant improvement over entropy-based approaches, with compression ratios ranging from approximately 2.7 to 9.9 across different datasets. Among them, LZMA consistently achieves the highest compression rates, particularly on structured datasets such as Clinical and Science, where it benefits from repetitive patterns. However, despite their improved efficiency, these methods still fall short of neural compression approaches in leveraging deeper contextual information.  

Neural-based compressors, including NNCP, TRACE, and PAC, further enhance compression performance, demonstrating that learned models can better capture the statistical structure of LLM-generated text. NNCP achieves particularly high compression rates on complex datasets such as Math and Science, while PAC and TRACE show competitive results across most categories. Notably, neural approaches tend to perform better on more structured datasets, suggesting that their learned representations effectively exploit latent regularities in LLM-generated content.  

The most striking result comes from the proposed method, which significantly outperforms all baseline compressors across every dataset. The compression rates achieved by our approach are consistently higher than those of both traditional and neural-based compressors, demonstrating its superior ability to model LLM-generated data. Notably, the largest improvements are observed on Science, Clinical, and Novel datasets, where compression rates surpass 20. This indicates that our method effectively captures and exploits the structured patterns inherent in LLM-generated text, leading to unprecedented gains in compression performance.  

\begin{figure*}[htbp]
    \centering
    \includegraphics[width=0.9\linewidth]{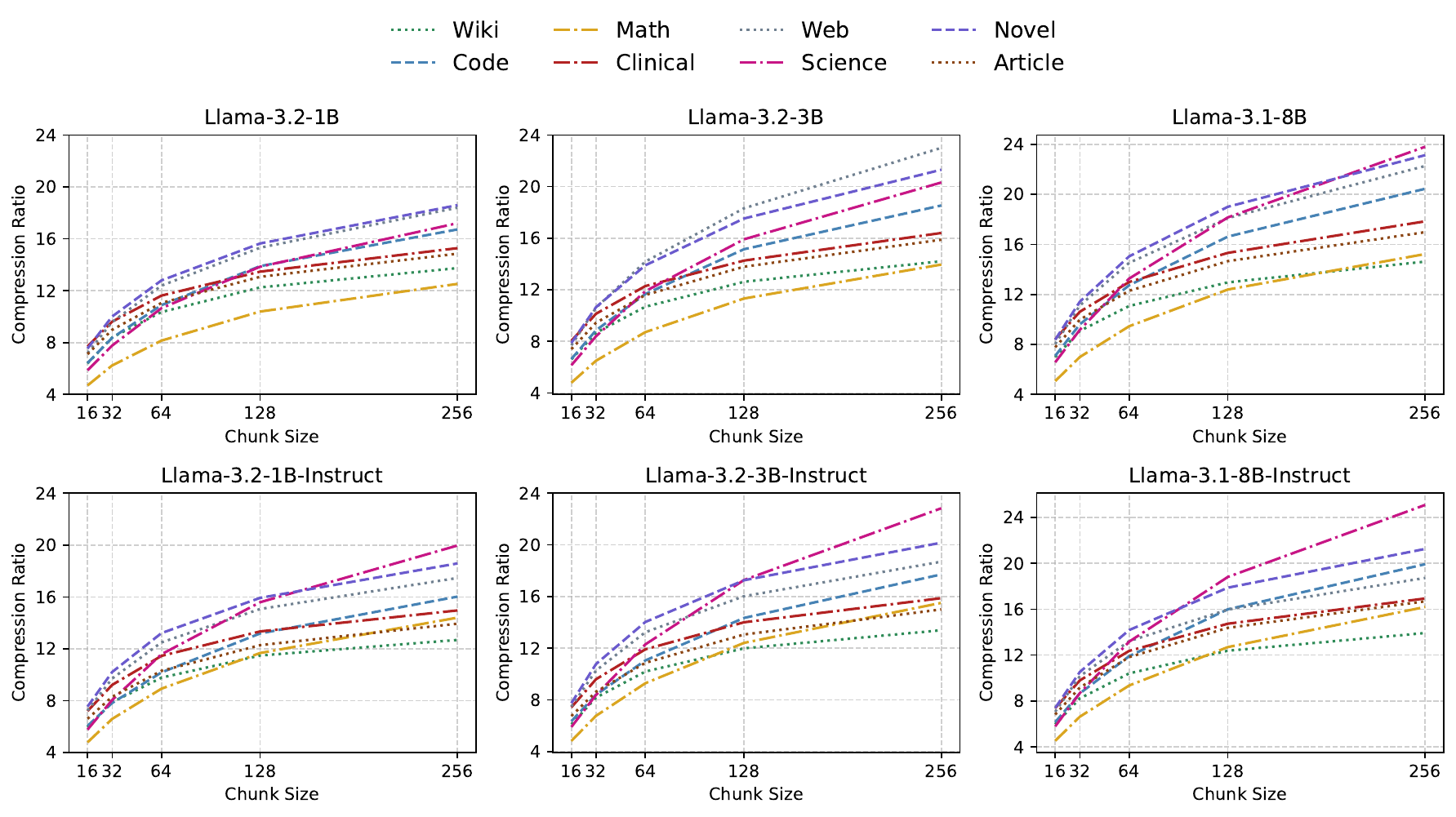} 
    \caption{Compression ratio comparison across different Llama models. Larger models achieve higher compression ratios, demonstrating the advantage of increased model capacity in leveraging contextual information. Instruction-tuned variants generally perform slightly worse than their base counterparts on most datasets but achieve better compression on question-answering datasets.}
    \label{fig:chunk_size_instruct} 
\end{figure*}

\subsection{Impact of Chunk Size on LLM Compression Peformance}

Chunk size significantly influences LLM compression performance by shaping how contextual information is utilized for token prediction. This subsection examines the effect of varying chunk sizes across different model configurations, including both base and instruction-tuned variants. 
Specifically, we test Llama-3.2-1B, Llama-3.2-3B, and Llama-3.1-8B, along with their instruction-tuned counterparts, across chunk sizes ranging from 16 to 256.

Experimental results for Llama-3.2-1B, Llama-3.2-3B, and Llama-3.1-8B, along with their instruction-tuned counterparts, consistently show that larger chunk sizes yield higher compression rates. This aligns with the theoretical premise that a broader context enables more accurate token predictions, reducing redundancy in the encoded representation. Compression efficiency improves progressively from chunk size 16 to 256, with larger models such as Llama-3.1-8B exhibiting steeper gains. Instruction-tuned models systematically outperform their base counterparts, particularly at larger chunk sizes, indicating that instruction tuning enhances token predictability. However, compression gains diminish beyond a certain threshold. The most notable improvements occur between chunk sizes 16 and 64, while gains taper off beyond 128, suggesting diminishing returns from additional context.

Although increasing the size of the fragment generally improves compression, the gain is not strictly linear. The most significant improvements are observed when transitioning from chunk size 16 to 64, whereas the relative improvement diminishes as chunk size increases beyond 128. This indicates that beyond a certain threshold, the additional context provided by larger chunk sizes yields diminishing returns in terms of compression performance.

\subsection{Impact of Model Size on Compression Peformance}
\begin{figure*}[htbp]
    \centering
    \includegraphics[width=0.85\linewidth]{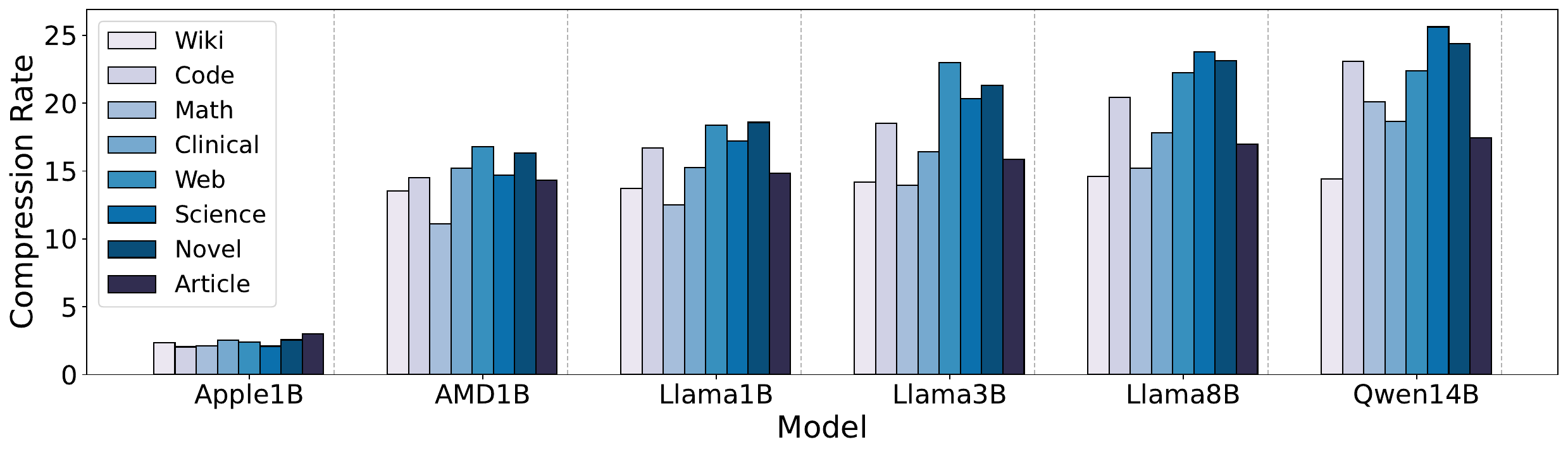} 
    \caption{Model scale vs. compression ratio.  
Larger models consistently demonstrate higher compression ratios, indicating that increasing model size enhances predictive capabilities and further reduces redundancy.}
    \label{fig:model_scale} 
\end{figure*}

One of the most straightforward yet effective ways to improve compression efficiency is to increase the model size. In this subsection, we analyze the impact of model size on compression rate by examining Fig.~\ref{fig:chunk_size_instruct}, which illustrates that larger models consistently achieve higher compression rates across both base and instruction-tuned variants. To provide a more detailed comparison, we further include Fig.~\ref{fig:model_scale}, which expands the analysis to six models: Apple1B, AMD1B, Llama1B, Llama3B, Llama8B, and Qwen14B. 

The experimental results reveal a clear upward trend in compression ratio with increasing model size. Across all conditions, the smallest models exhibit the lowest compression rates, while the largest models, such as Llama8B and Qwen14B, demonstrate significantly higher compression performance. This pattern holds across different datasets and configurations, reinforcing the notion that larger models benefit from enhanced predictive capabilities, which allow them to reduce redundancy more effectively. Due to hardware constraints, our experiments are limited to models up to 14B parameters. We believe that 14B is sufficient to cover most LLM compression applications, as it already represents a large-scale model. However, we acknowledge that exploring even larger models remains an interesting direction for future research.

\subsection{Impact of Dataset Scale on Compression Performance}

Dataset scale is another crucial factor influencing compression efficiency. In this subsection, we investigate whether increasing the size of LLM-generated synthetic data affects compression performance. To this end, we evaluate both traditional and LLM-based compressors under different scales of the Wiki dataset.

\begin{figure}[ht]
    \centering
    \includegraphics[width=0.8\linewidth]{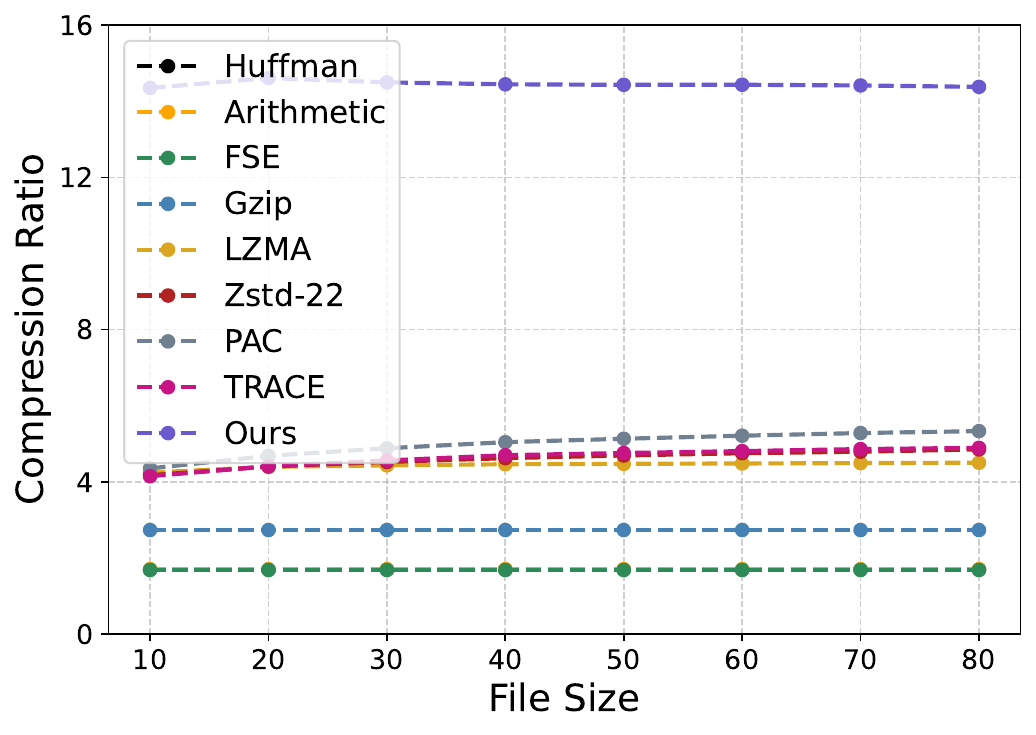} 
    \caption{Compression ratio variations with increasing dataset size. Traditional methods remain stable, while neural compressors show slight improvement, but the proposed method consistently outperforms all approaches.}
    \label{fig:file_scale} 
\end{figure}

The results are given in Fig.~\ref{fig:file_scale}.
Traditional methods such as Huffman coding, Arithmetic coding, FSE coding, and Gzip exhibit no significant change in compression rate, indicating that their underlying encoding mechanisms do not benefit from increased data availability. More advanced methods like LZMA and Zstd-22 show a slight improvement in compression rate with larger datasets, but the growth remains modest. In contrast, NN-based compressors such as TRACE and PAC demonstrate both superior compression performance and a slightly higher rate of improvement as dataset size increases. However, none of these methods come close to matching the compression efficiency achieved by the proposed approach.  

A particular observation is that the proposed method maintains a consistent compression rate regardless of dataset scale. The stability of the proposed method across different dataset scales suggests potential applications in scenarios where data volume fluctuates, such as real-time streaming or incremental data storage.

\subsection{Comparison between LLMs with and without Fine-tuning}

\subsubsection{Instruction tuning}

Instruction tuning is an indispensable post-training procedure for existing deployed LLMs, which aims to enable LLMs to align with human preference data and not exhibit unsafe behaviors.
As instruction tuning also fine-tunes the token distributions of the base LLMs,
in this subsection, we conduct an investigation into the influence of instruction tuning on the compression capabilities of LLMs for the text data.

The experimental results reveal that instruction-tuned variants generally exhibit a slightly lower compression ratio compared to their base model across most datasets. This phenomenon can be attributed to the differences in training objectives between pre-training and instruction tuning. 
During the initial training phase of large language models, the objective function is designed to minimize the cross-entropy loss between the predicted token distribution and the true labels, which aligns naturally with the goal of compression (minimizing entropy). However, instruction tuning introduces a different optimization objective that alters the token distribution, potentially reducing its effectiveness for compression.  

Unlike pretraining, instruction tuning focuses on aligning model outputs with human preferences rather than purely maximizing predictive accuracy. This process involves reinforcement learning with human feedback (RLHF) or supervised fine-tuning on instruction-following datasets, where the model is explicitly optimized to generate responses that are coherent, contextually relevant, and stylistically appropriate rather than merely predicting the most probable next token. These factors can lead to a token distribution that is less compressible, as the model is encouraged to generate structured, human-like responses rather than the statistically optimal token sequences for entropy minimization.  

One example supporting this hypothesis is the observation that instruction-tuned models achieve a consistent increase in compression efficiency on science and math datasets across different model sizes. Since these datasets primarily consist of structured question-answer pairs, their underlying probability distributions are more aligned with the instruction-tuning objective, which emphasizes coherent and structured responses. This alignment likely results in improved token predictability within these domains, leading to better compression rates.  

\subsubsection{Specific Domain Fine-tuning}

To verify the compressibility of domain-specific models on their respective domains, we evaluate compression ratios using specialized models on math and code datasets. For the math dataset, we select Qwen2.5-Math-1.5B and Rho-Math-1B. For the code dataset, we use Qwen2.5-Coder-1.5B and DeepSeek-Coder-1.3B-Instruct. To ensure fairness, all selected models have sizes ranging from approximately 1B to 1.5B.

\begin{figure}[t]
    \centering
    \includegraphics[width=0.95\linewidth]{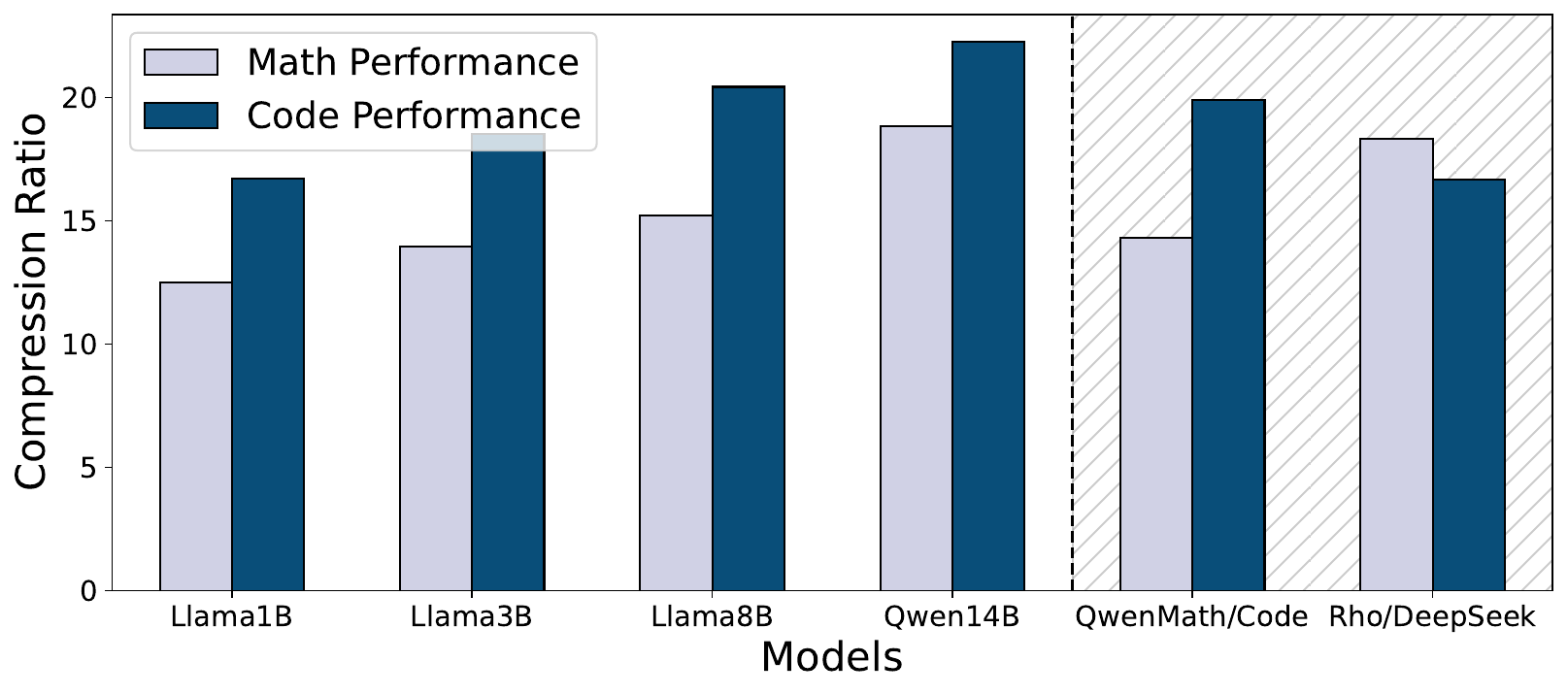} 
    \caption{Performances of domain-specific models on Math and Code tasks. Domain-specific models consistently achieve higher compression ratios within their respective domains.}
    \label{fig:domain} 
\end{figure}

Table~\ref{fig:domain} presents the compression performance of general models and two domain-specific models on math and code tasks. While larger models such as Qwen14B and Llama8B generally achieve higher compression ratios, the domain-specific models exhibit significantly better compression efficiency.
On the math dataset, when compared to similarly sized 1B models, Qwen2.5-Math-1.5B and Rho-Math-1B achieve noticeably higher compression ratios. Remarkably, Rho-Math-1B reaches a compression ratio comparable to Qwen14B, despite being 14 times smaller in size. 
For the code dataset, Qwen2.5-Coder-1.5B attains a compression ratio similar to Llama8B. However, DeepSeek-Coder-1.3B performs comparably to the general-purpose Llama1B model, which we suspect may be due to insufficient training.
This suggests that models explicitly trained or fine-tuned for specific domains can better capture the structural patterns and redundancies inherent in domain-specific data, leading to improved compression rates.

\subsection{Predictability of LLMs on Human vs. LLM-Generated Data}
In this subsection, we study the predictability of LLMs for human-generated data and LLM-generated data. We evaluate the compression ratios of LLMs under different chunk sizes for the two categories of text data. Specifically, we run Llama-3.2-1B on real movie critics extracted from imdb and LLM-generated movie critics.

\begin{figure}[t]
    \centering
    \includegraphics[width=0.8\linewidth]{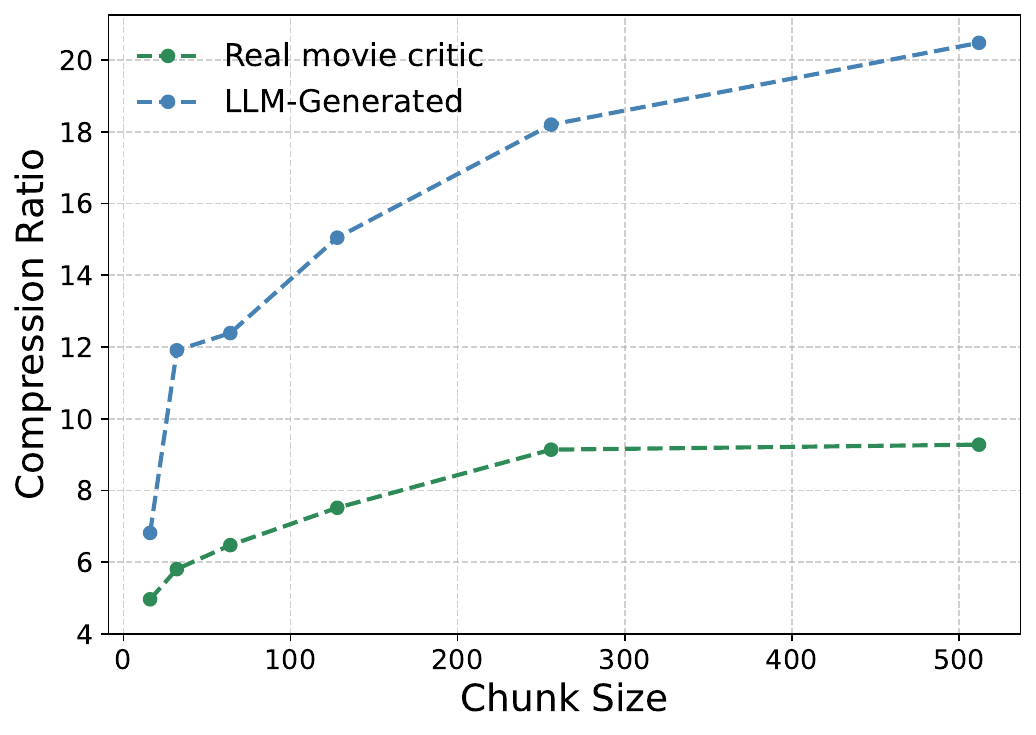} 
    \caption{Compression ratio comparison between human-generated and LLM-generated text. LLMs achieve significantly higher compression ratios on LLM-generated data, with the gap widening as chunk size increases.}
    \label{fig:chunk_human} 
\end{figure}

The results are shown in Figure~\ref{fig:chunk_human}. We can find that under different chunk sizes, LLMs demonstrate significantly higher compression ratios on LLM-generated data over human data. More interestingly, along with the increase in the chunk size, the compression ratios increase even further. Therefore, in large-scale data compression, as the ratios of LLM-generated data keep growing, LLM-based compressors will demonstrate increasingly significant compression capabilities compared to traditional compressors.

\section{Related Work}
\subsubsection{Traditional Compression.}
Traditional lossless compression methods leverage entropy coding and dictionary-based techniques to efficiently reduce data size while preserving exact information. Huffman coding~\cite{huffman1952method} assigns variable-length codes based on symbol frequency, ensuring shorter codes for more common symbols. Arithmetic coding~\cite{arithmetic} provides better compression by representing entire sequences as fractional numbers, achieving near-optimal entropy efficiency. Dictionary-based methods like Gzip~\cite{gzip} and LZMA~\cite{lz77} utilize Lempel-Ziv compression, replacing repeated substrings with references to previous occurrences. Zstd~\cite{zstd}, an advanced variant, improves upon Gzip with faster compression and decompression speeds while maintaining high compression ratios.

\subsubsection{Neural Network-based Compressors}
The use of neural networks as predictors for lossless compression has been widely explored, particularly in conjunction with arithmetic coding~\cite{Cox2016,Goyal2019,bellard2023nncp}.
NNCP~\cite{bellard2023nncp} employs transformers to estimate probability distributions, while TRACE~\cite{mao2022trace} introduces a lightweight transformer variant to accelerate the compression process. Some works start using foundation models for lossless compression recently~\cite{Deletang2024, Huang2024, Mittu2024}. Studies leveraging models such as LLaMA\cite{llama3.2} and Chinchilla~\cite{Deletang2024} demonstrate promising results, particularly in compressing previously unseen modalities—for instance, large language models trained on text effectively compress images and audio. However, none of these studies specifically address \textit{the domain of synthetic data}, leaving a gap in exploring how such models perform in artificial data compression scenarios.

\section{Conclusion}

As LLM-generated data continues to grow exponentially, effective lossless compression of LLM-generated data has become a critical challenge in text data management. In this work, we conducted the first systematic investigation into compressing LLM-generated text.
We demonstrated that traditional lossless compressors struggle to effectively compress LLM-generated data due to the irregularity and complexity of LLM-generated data.
Unlike previous machine-generated data, LLM-generated data lacks structured redundancy and appears more similar to human-generated data, posing critical challenges to lossless compression.
To mitigate the issue, we proposed leveraging the predictive capabilities of LLMs originating from next-token prediction. Our extensive experiments with 14 LLMs and 8 datasets revealed that LLM-based prediction methods achieve over 20× compression, far surpassing Gzip’s 3× compression. Furthermore, we show that this approach generalizes across different model scales and dataset categories, underscoring the robustness and versatility of LLM-based compression. 

\bibliographystyle{ACM-Reference-Format}
\bibliography{sample-base}
\end{document}